\documentclass[11pt, a4paper]{article}
\usepackage{coling2020}
\usepackage{times}
\usepackage{graphicx}
\usepackage{tabularx}
\usepackage{soul}

\usepackage{epstopdf}
\usepackage[latin1]{inputenc}

\usepackage{hyperref}
\usepackage{xstring}

\usepackage{latexsym}
\usepackage{url}
\usepackage{dsfont}
\usepackage{textcomp}
\usepackage{amsfonts}
\usepackage{amsmath}
\usepackage{amssymb}
\usepackage{dirtytalk}
\usepackage{graphicx}
\usepackage{booktabs}
\usepackage{multirow}
\usepackage[dvipsnames]{xcolor}

\usepackage{etoolbox} 

\usepackage{array}
\newcolumntype{H}{>{\setbox0=\hbox\bgroup}c<{\egroup}@{}}

\newcommand{\dsfullname}{New York Times Word Innovation Types}
\newcommand{\dsname}{NYTWIT}
\newcommand{\labl}[1]{\textsc{#1}}

\newcommand{\clj}[1]{\textcolor{magenta}{[Cassie: \say{#1}]}}

\title{\dsname: A Dataset of Novel Words in the New York Times}

\makeatletter
\newcommand{\printfnsymbol}[1]{%
  \textsuperscript{\@fnsymbol{#1}}%
}
\makeatother

\author{Yuval Pinter \\
  Georgia Institute of Technology \\
  Atlanta, GA, USA \\
  {\tt uvp@gatech.edu} \\\And
  Cassandra L. Jacobs \\
  University of Wisconsin \\
  Madison, WA, USA \\
  {\tt cjacobs2@wisc.edu} \\\And
  Max Bittker \\
  School for Poetic Computation \\
  New York, NY, USA \\
  {\tt maxbittker@gmail.com} \\}

\date{}

\begin{document}
\maketitle

\begin{abstract}
    We present the \dsfullname{} dataset, or \textbf{\dsname}, a collection of over 2,500 novel English words published in the New York Times between November 2017 and March 2019, manually annotated for their class of novelty (such as lexical derivation, dialectal variation, blending, or compounding). 
    We present baseline results for both uncontextual and contextual prediction of novelty class, showing that there is room for improvement even for state-of-the-art NLP systems.
    We hope this resource will prove useful for linguists and NLP practitioners by providing a real-world environment of novel word appearance.
\end{abstract}

\section{Introduction}
\label{sec:intro}

Novel words, or Out-Of-Vocabulary words (OOVs), are a pervasive problem in modern natural language processing \cite{brill1995transformation,young2018recent}.
A common scenario in which this problem appears is that of a pre-trained model containing a word representation component such as an embedding table, encountering a previously-unseen word in a downstream task such as question answering or natural language inference.
Multiple lines of work attempt to alleviate the downstream effect of OOV words \cite{mueller2011improved,mimick}, but most tend to focus on individual categories of OOVs: typographical errors \cite{sakaguchi2017robsut}, domain-specific terminology \cite{du2016using}, stylistic variability \cite{eisenstein2013bad,vandergoot2019depth}, morphological productivity \cite{bhatia2016morphological}, or novel named entities \cite{hoffart2014discovering}.
In reality, unseen texts contain all these classes of novelty, and more.
OOVs are a typically presented as a significant challenge for generalization or understanding in noisy user-generated text (e.g.~Twitter) and/or domain-specific content.
Nevertheless, even large corpora that are narrow in domain (edited news stories) contain linguistic innovations, including but not limited to novel morphological processes, typographical errors, and loan words. 


In this paper, we present a dataset of novel words in English relative to the corpus of articles published by the New York Times (NYT), as collected automatically in real time by a Twitter bot.
We name it the \dsfullname{} corpus, or \textbf{\dsname} for short.
We annotated each word for one of eighteen linguistically-informed \textbf{categories} of novelty within the context of the NYT corpus, as well as for its date of publication and a retrieval document identifier to enable context extraction.\footnote{Context article excerpts are not freely available without copyright licensing from the New York Times, who have ignored all contact attempts to date.}
To our knowledge, this is the first resource to include novel words along with their contextual information in addition to linguistically-informed annotation, a method that enables expansion beyond dictionary-based methods \cite{cook-stevenson-2010-automatically,dhuliawala-etal-2016-slangnet,ahmad2000neologisms} and decontextualized neologisms \cite{kulkarni-wang-2018-simple}.
In contrast with resources which provide examples and attestations to lexical forms, \dsname{} was constructed in a corpus-comprehensive manner where novelty guides curation and not vice versa.

In addition, we provide results for the task of classifying words into their categories based on word form and contextual information, a task which can both provide data for linguistic analysis of lexical enrichment and serve as a processing step for NLP systems which may work better if different modules are applied to different classes of novel words.
We show that both character-level models and large pre-trained sentence encoders struggle on this task, illustrating the challenges of modeling language innovation.

We release the data at \url{https://github.com/yuvalpinter/nytwit} under the GNU General Public License v3.0.
The project is ongoing, and this document pertains to version 1.1.

\section{The \dsfullname{} Dataset}

Our dataset is built upon two bots developed by the third author.
The first stage of data collection relies on tweets from the NYT\_First\_Said bot\footnote{\url{https://twitter.com/NYT\_First\_Said}}, which operates by scraping new articles as they post on the NYT site and tweeting out novel words following a filtering process which we will describe at a high level.\footnote{The code for the bot is available at \url{https://github.com/MaxBittker/NYT-first-said}.}
After tokenizing on white space and punctuation, the precision-oriented script rejects capitalized words in order to avoid proper nouns (at the cost of missing sentence-initial true OOVs).
\texttt{langid} \cite{lui2012langid} is used to reject non-English sentences, while still allowing loanwords in English sentences.
Words are queried against the historical NYT search API to detect unpublished words.\footnote{We note that the search index relies on imperfect, although extensive, digitization artifacts.
At the time of writing, in a sample of 450 terms from our dataset, \clj{4} were entries in the Oxford English Dictionary, nearly all of which belong to the domain or foreign categories.}
For the time range of our collected corpus, November 7, 2017 to March 28, 2019, a bandwidth limit of five words per 30 minutes was imposed, but we confirmed that this did not have a substantial effect on OOV coverage, leaving our artifacts distributionally representative for the news domain.

An associated context bot replies to the tweets with links to the original articles.\footnote{\url{https://twitter.com/NYT\_Said\_Where}}
We used the URLs from this bot's posts as the main reference for the words' contexts.
For 17 words, the article URL was retrieved manually by searching for the target article directly.\footnote{One term lacks a context because neither the NYT search engine nor the API support the letter \'e.}
As the articles are subject to edits long after publication, there is an increasing but small portion of articles which no longer contain the context, although at time of publication these mostly include the removal of typographical errors from the stories and which are ultimately filtered by our annotation process (see below).

\subsection{Annotation}
\label{sec:ann}

The extracted data was independently annotated and filtered by the first two authors.
Initially, all 2,587 words were assigned one of 20+ tags inspired by the word formation literature \cite{kiparsky1982word,klymenko2019twitterverse}.
Certain categories were filter categories intended to capture and exclude false positives from the final dataset: \labl{Duplicate} for inflections of words already appearing in the dataset in a morphologically simpler form, e.g.~\textit{batchcode} and \textit{batchcodes};
\labl{Foreign} and \labl{PRP} for foreign words and proper names (mostly all-lowercase Twitter usernames) which were not caught by the automatic filtering;
\labl{Spaces} and \labl{Typo} for unintended cases of space deletion and typographical errors which were not caught by NYT editors.\footnote{The overwhelming share of these words have indeed since been deleted from the NYT website.}
The filtered items are provided in the dataset under the label \labl{Filtered}.

Agreement between the annotators at the preliminary phase was 68\% over all labels, and 0.65 Cohen's Kappa.
Following category filtering, amounting to 40\% of the original dataset, agreement over the remaining 1,550 words was calculated to be 65\% at 0.61 Kappa.  
At the coarse-grained level, agreement on the four themes (lexical / morphological / syntactic / sociopragmatic) was 89\% at 0.75 Kappa.\footnote{A reviewer noted that these are low agreement rates, and compared the task to part-of-speech annotation. We dispute the comparison, both on grounds of the novelty of the forms involved and of the mechanical syntactic nature of the majority of POS tagging decisions.}

The annotators then examined each other's annotations and agreed on some consolidation of rarely-occurring original labels, as well as introduction of new labels deemed useful post-hoc.

\subsection{Novel Word Taxonomy}
We describe the eighteen categories in the finalized dataset, organized by a thematic grouping not explicitly annotated.
Counts for each category are provided [in brackets].

\paragraph{Lexical OOVs.}
We deem certain categories to arise from the fact that the NYT, while being interested in many aspects of life, has not had the chance to delve into each and every one at depth over its 168 years of existence.
These are the \labl{Domain} label for technical terms from uncommon domains (e.g.~\textit{glossopoeia}) [285];
the \labl{innovation} label for terms coined with no discernable prevailing linguistic process (e.g.~\textit{swanicles}, a term from a work of fiction) [11];
and the \labl{Onomatopeia} label for sound-based sequences (e.g.~\textit{ktktk}) [23], which includes cases of verbatim vocalization such as \textit{trololo}.

\paragraph{Morphological OOVs.}
In this group we include categories of words composed of meaning-carrying units present in existing English words which have appeared in the NYT before, manifested in a new form.
In increasing order of syntactic and semantic novelty, they are:
\labl{Infl}, unseen inflections of existing wordforms: same part-of-speech, different syntactic attributes (e.g.~\textit{pennyloafers}) [53];\footnote{We include the negating prefixes \texttt{in-} and \texttt{un-}, which despite change a word's meaning, but retain its part-of-speech.}
\labl{Deriv}, unseen derivations of existing words into new parts-of-speech which carry no semantic distancing beyond that implicit in the new part-of-speech itself (e.g.~\textit{foamability}) [215];
\labl{Affix}, affixation of very distinct base words which are typically derivational in nature but include a semantic charge (e.g.~\textit{extraphotographic}, \textit{pizzaless}) [483];
\labl{Affix\_Libfix}, affixation of distinct base words with particles extracted from another word in a process known as \textit{libfixation} \cite{zwicky2010libfixes} or \textit{splintering} \cite{berman1961contribution}, where the liberated affix still elicits the originating word but can be freely attached to a growing selection of words (e.g.~\textit{dripware}) [18];
\labl{Compound\_Comp}, a concatenation of two complete words each contributing essential semantics to the final form in a way we deem (subjectively, with help of context) to be compositional (e.g.~\textit{smellwalks}, strolls focusing on olfactory input) [121];\footnote{One compound in our dataset, \textit{dramatotherapy}, adds characters for cadence; another, \textit{laysoccerperson}, is nonlinear.}
\labl{Compound\_New}, a concatenation of base words resulting in a new semantic concept deemed remote from the bases (e.g.~\textit{nothingbuffet}, a play on \textit{nothingburger}) [49]; and
\labl{Blend}, a fusion of two or more base forms together where original characters are lost or shared, or new ones are added (e.g.~\textit{chipster}, a chicano hipster) [142].\footnote{A single blend, \textit{pregret}, has just one base fused with a prefix.}

\paragraph{Syntactic OOVs.}
This group consists solely of the \labl{Synth} category of tokens which synthesize multiple syntactic words into one form, a rare formation process in English limited typically to auxiliary contractions (e.g.~\textit{this'll}) [6].

\paragraph{Sociopragmatic OOVs.}
Words in this group exhibit an orthographic diversion from standard English usually intended as a statement of register or status, or as a faithful representation of a certain linguistic style or sentiment.
\labl{Archaic}, a register of older variants of English or an ironic semblance of such (e.g.~\textit{shooketh}, a mock-archaic form of \textit{shake} using Middle English morphology) [14];
\labl{Dialect}, a geographically- or demographically-specific form of a word typically spelled differently in the NYT (e.g.~\textit{skwarsh}, an r-full \textit{squash}) [46];
\labl{Infix}, a morphological tool reserved in English for expletive emphasis \cite{mccawley1978you} (e.g.~\textit{unfreakingbelievable}) [2];
\labl{Phonaestheme}, a phonological duplication phenomenon used in contemporary English nearly only as derisive echo reduplication borrowed from Yiddish \cite{wales1990phonotactics} (e.g.~\textit{schmarket}) [6];
\labl{Lengthening}, a written manifestation of the expressive elongation of phonetic segments (e.g.~\textit{greaaaaat}) [53];
\labl{Variant}, spelling alternations or intentional typos which are not intended to be read differently from the standard form of the word, used for branding and jest (e.g.~\textit{kyllyng}) [18]; and
\labl{Spaces\_Sic}, the removal of whitespace to simulate breathlessness (e.g.~\textit{lineafterlineafterline}) [5].

\subsubsection{Difficult Distinctions}

Naturally, some annotation cases are not clear-cut, as evidenced by the imperfect inter-annotator agreement.
We found the most challenging cases to be among the morphological categories, where an affix is either semantically null (\labl{Deriv} / \labl{Infl}) or not (\labl{Affix}) (14\% and 15\% of disagreements, respectively); where a sense of the nearest in-vocabulary word can signal the difference between \labl{Infl} and \labl{Deriv} (3.4\%);
where an \labl{Affix\_Libfix} has been \say{liberated} enough from the underlying word such that it is now simply an \labl{Affix} (does \textit{cyber-} still envoke the full word \textit{cybernetics}? Does \textit{crypto-} envoke \textit{cryptography}?);
if it has not been liberated yet, it should be a \labl{Blend} or a \labl{Compound}.
In addition, the pre-processing phase required a demarcation between \labl{Domain} and \labl{Foreign} which was not easy to make given the heavy foreign-word influence in certain knowledge domains such as cuisine (e.g.~\textit{dinkelbrot}).
Words adapted into English morphology would usually lead to a \labl{Domain} label (\labl{Domain} vs. \labl{Compound}: 4\%).
In many cases, we found the contexts in which the words were introduced to give sufficient disambiguation (so, e.g.,~\textit{cybercoach} is an affix, but \textit{cyberinvasion} is a compound).

We invite readers to email errata to either of the first two authors, or submit a pull request on Github.

\section{OOV Classification Task}

The task of classifying OOVs, i.e. assigning a novel word with a label from the taxonomy we defined above, can be beneficial from both an analytical linguistic standpoint, and from an NLP standpoint concerned with model performance on downstream language understanding tasks.
To get a sense of the predictability of the various OOV classes in the dataset, we present several baselines for this straightforward task.
The uniqueness of our dataset allows us to apply both type-level and context-dependent systems, the latter operating in the real-world scenario of encountering a word for the first time in the actual context of its introduction to the corpus. 

First, our \textbf{Majority class} baseline assumes all OOVs are the result of \textit{affixation}.

For all following models we trained a ridge classifier with default regularization parameters in \texttt{scikit-learn}. 
Scores for all supervised models are reported via 10-fold cross-validation using the same folds for all systems.
Due to the class imbalance, we chose to implement training in such a way that upsampled rare classes with replacement at each iteration to equal frequency as the most common class. We report accuracy (\textsc{Acc}) and macro F1 scores.

\paragraph{Contextless features.}
We compare and contrast several input features to our classifier that only have access to the form of the OOV, without consideration of the context:
\begin{itemize}
    \item \textbf{Character n-grams.}
    We extract bag-of-character features ranging from one to three characters for each OOV.
    The feature vocabulary is estimated on the training set and applied to the test set.
    \item \textbf{FastText.} We infer fasttext vectors~\cite{fasttext}, applying its 3--6 character-ngram representations, from the subword model trained on English Wikipedia.\footnote{\texttt{wiki.en.bin} file obtained May 25, 2020.}
    \item \textbf{ELMo embeddings.} We use the word-level embeddings from ELMo \cite{elmo}, obtained via a pre-trained character-level convolutional net for each OOV presented in isolation, with no surrounding sentence context.
    \item \textbf{BERT no-context.} We apply BERT-Base~\cite{devlin2019bert} to the OOVs in a null \say{\texttt{[CLS] \_\_\_ [SEP] .}} context.
    The averaged top-layer vectors from all the OOV's word pieces are passed to the classifier.\footnote{Using just the embedding of the final word piece produced similar results.}
\end{itemize}

\paragraph{Context-aware features.}
\begin{itemize}
    \item \textbf{Character RNN.}
    We train a 2-layer forward- (backward-) character-level GRU language model on 100,000 Wikipedia documents and run it through the beginning (end) of the sentence until the OOV, then use the concatenated final hidden states from each direction as features.
    \item \textbf{ELMo.} We obtain contextualized embeddings for all words in our sentences and select the top layer representation associated with each OOV.
    \item \textbf{BERT.} We apply BERT-Base to the entire sentence in which the OOV appears, and use the averaged top-layer embeddings at the indices of each OOV.
\end{itemize}

\subsection{Results}
The results, presented in Table~\ref{tab:results}, show that pretrained contextual models not only trail behind a contextless, un-pretrained character n-gram baseline, they even fail to improve over their own uncontextualized variants.
An analysis of class-specific F1 scores across the different models exposed two general patterns in classifier performance:
in all models, performance on the \labl{Affix} class was in the top four, and the same for \labl{Lengthening} except for Character RNN.
We also observed that models that encode contextual, sentence-level properties are typically better at encoding genre phenomena (e.g.~\labl{Domain} was a top-four category for BERT, Character RNN, fastText, and ELMo).
However, for some classes of models, there was a clear benefit to memorizing word forms.
All count-based feature representations (e.g.~bag-of-character ngrams, bag-of-wordpieces) led to better performance on orthographic properties, namely \labl{Phonaestheme}, \labl{Synth}, and \labl{Onomatopoiea}.
These results demonstrate the power that simple surface-form signals from character sequences still possess in meaningful NLP tasks.
In future work, we will attempt to supplement the contextual models with auxiliary mechanisms and perform fine-tuning.

\begin{table}
    \centering
    \small
    \begin{tabular}{lcHc|lcHc}
        \toprule
        Contextless & \textsc{Acc} & \textsc{mic} & \textsc{F1} & Contextual & \textsc{Acc} & \textsc{mic} & \textsc{F1} \\
        \midrule
         Majority class & .312 & & .026 \\
         Character n-grams & \textbf{.484} & & \textbf{.323} \\
         FastText & .433 &  & .241 & Character RNN & .128 & & .054 \\
         ELMo embeddings & .365 & & .203 & ELMo & .324 & & .135 \\
         BERT no-context & .442 & & .288 & BERT & .469 & & .269 \\
        \bottomrule
    \end{tabular}
    \caption{Baseline results for OOV classification ($N=1550$, $|C|=18$).}
    \label{tab:results}
\end{table}

\section{Conclusion}
\label{sec:conc}

We presented a novel dataset of OOVs along with their contexts and linguistic novelty class annotations.
We showed that contextual information in the form of other parts of the sentence provides some signal, but simple models relying on character n-gram information alone achieve high performance.

The availability of broader document contexts in which these neologisms occur enables many linguistic and technical applications.
From the perspective of the study of language growth and formation, the dataset may be used to assess the morphological productivity of different affixes and roots, or the prevalence of the different word formation processes in a realistic setting; or perform in-depth analysis on any of the specific types of innovations we identified.
In addition, the in-vivo nature of the dataset provides a reference for neologisms which may or may not be later adopted into everyday use, allowing diachronic studies anchored in the time of word introduction.
Analysis of the phonological, morphological, and discourse-level properties of these words may provide insight into lexical adoption dynamics.

For NLP researchers, an important component of text applications is proper normalization and segmentation of word forms. 
Our experiment shows that popular word form encoders, such as ELMo or BERT's WordPiece, still have a long way to go in terms of recognizing the origins of a novel form.
Errors at this stage might lead to inability to handle morphologically complex OOVs in downstream semantic applications~\cite{unblend}, although further study of such effects and of the utility of OOV classification in alleviating them is still necessary.
Properly leveraging context for morphological decomposition of complex forms also remains an open problem.

The resource is an ongoing project; the repository includes plans for the next versions, including increasing the dataset size by including newer words from the bot, and annotating additional information such as part-of-speech tags.

\section*{Acknowledgments}

We thank Jacob Eisenstein, Kyle Gorman, Arya McCarthy, Sandeep Soni, and the anonymous reviewers for their valuable notes.
Yuval Pinter is a Bloomberg Data Science PhD Fellow.
Cassandra Jacobs is supported on NSF BCS Grant 1849236 awarded to Maryellen MacDonald.

\bibliographystyle{acl}
\bibliography{nyt-first-said}

\begin{thebibliography}{}

\bibitem[\protect\citename{Ahmad}2000]{ahmad2000neologisms}
Khurshid Ahmad.
\newblock 2000.
\newblock Neologisms, nonces and word formation.
\newblock In {\em Proceedings of the Ninth EURALEX International Congress},
  pages 711--730.

\bibitem[\protect\citename{Berman}1961]{berman1961contribution}
JM~Berman.
\newblock 1961.
\newblock Contribution on blending.
\newblock {\em Zeitschrift f{\"u}r Anglistik und Amerikanistik}, 9:278--281.

\bibitem[\protect\citename{Bhatia \bgroup et al.\egroup
  }2016]{bhatia2016morphological}
Parminder Bhatia, Robert Guthrie, and Jacob Eisenstein.
\newblock 2016.
\newblock Morphological priors for probabilistic neural word embeddings.
\newblock In {\em Proceedings of the 2016 Conference on Empirical Methods in
  Natural Language Processing}, pages 490--500, Austin, Texas, November.
  Association for Computational Linguistics.

\bibitem[\protect\citename{Bojanowski \bgroup et al.\egroup }2017]{fasttext}
Piotr Bojanowski, Edouard Grave, Armand Joulin, and Tomas Mikolov.
\newblock 2017.
\newblock Enriching word vectors with subword information.
\newblock {\em Transactions of the Association for Computational Linguistics},
  5:135--146.

\bibitem[\protect\citename{Brill}1995]{brill1995transformation}
Eric Brill.
\newblock 1995.
\newblock Transformation-based error-driven learning and natural language
  processing: A case study in part-of-speech tagging.
\newblock {\em Computational Linguistics}, 21(4):543--565.

\bibitem[\protect\citename{Cook and
  Stevenson}2010]{cook-stevenson-2010-automatically}
Paul Cook and Suzanne Stevenson.
\newblock 2010.
\newblock Automatically identifying the source words of lexical blends in
  {E}nglish.
\newblock {\em Computational Linguistics}, 36(1):129--149.

\bibitem[\protect\citename{Devlin \bgroup et al.\egroup }2019]{devlin2019bert}
Jacob Devlin, Ming-Wei Chang, Kenton Lee, and Kristina Toutanova.
\newblock 2019.
\newblock Bert: Pre-training of deep bidirectional transformers for language
  understanding.
\newblock In {\em Proceedings of the 2019 Conference of the North American
  Chapter of the Association for Computational Linguistics: Human Language
  Technologies, Volume 1 (Long and Short Papers)}, pages 4171--4186.

\bibitem[\protect\citename{Dhuliawala \bgroup et al.\egroup
  }2016]{dhuliawala-etal-2016-slangnet}
Shehzaad Dhuliawala, Diptesh Kanojia, and Pushpak Bhattacharyya.
\newblock 2016.
\newblock {S}lang{N}et: A {W}ord{N}et like resource for {E}nglish slang.
\newblock In {\em Proceedings of the Tenth International Conference on Language
  Resources and Evaluation ({LREC}'16)}, pages 4329--4332, Portoro{\v{z}},
  Slovenia, May. European Language Resources Association (ELRA).

\bibitem[\protect\citename{Du \bgroup et al.\egroup }2016]{du2016using}
Jinhua Du, Andy Way, and Andrzej Zydron.
\newblock 2016.
\newblock Using {B}abel{N}et to improve {OOV} coverage in {SMT}.
\newblock In {\em Proceedings of the Tenth International Conference on Language
  Resources and Evaluation ({LREC}'16)}, pages 9--15, Portoro{\v{z}}, Slovenia,
  May. European Language Resources Association (ELRA).

\bibitem[\protect\citename{Eisenstein}2013]{eisenstein2013bad}
Jacob Eisenstein.
\newblock 2013.
\newblock What to do about bad language on the internet.
\newblock In {\em Proceedings of the 2013 Conference of the North {A}merican
  Chapter of the Association for Computational Linguistics: Human Language
  Technologies}, pages 359--369, Atlanta, Georgia, June. Association for
  Computational Linguistics.

\bibitem[\protect\citename{Hoffart \bgroup et al.\egroup
  }2014]{hoffart2014discovering}
Johannes Hoffart, Yasemin Altun, and Gerhard Weikum.
\newblock 2014.
\newblock Discovering emerging entities with ambiguous names.
\newblock In {\em Proceedings of the 23rd international conference on World
  wide web}, pages 385--396.

\bibitem[\protect\citename{Kiparsky}1982]{kiparsky1982word}
Paul Kiparsky.
\newblock 1982.
\newblock Word-formation and the lexicon.
\newblock In {\em Proceedings of the Mid-America Linguistics Conference}, pages
  3--29. University of Kansas.

\bibitem[\protect\citename{Klymenko}2019]{klymenko2019twitterverse}
Olga Klymenko.
\newblock 2019.
\newblock Twitterverse: The birth of new words.
\newblock {\em Proceedings of the Linguistic Society of America}, 4(1):11--1.

\bibitem[\protect\citename{Kulkarni and Wang}2018]{kulkarni-wang-2018-simple}
Vivek Kulkarni and William~Yang Wang.
\newblock 2018.
\newblock Simple models for word formation in slang.
\newblock In {\em Proceedings of the 2018 Conference of the North {A}merican
  Chapter of the Association for Computational Linguistics: Human Language
  Technologies, Volume 1 (Long Papers)}, pages 1424--1434, New Orleans,
  Louisiana, June. Association for Computational Linguistics.

\bibitem[\protect\citename{Lui and Baldwin}2012]{lui2012langid}
Marco Lui and Timothy Baldwin.
\newblock 2012.
\newblock langid. py: An off-the-shelf language identification tool.
\newblock In {\em Proceedings of the ACL 2012 system demonstrations}, pages
  25--30. Association for Computational Linguistics.

\bibitem[\protect\citename{McCawley}1978]{mccawley1978you}
James~D McCawley.
\newblock 1978.
\newblock Where you can shove infixes.
\newblock {\em Syllables and segments}, pages 213--221.

\bibitem[\protect\citename{M{\"{u}}ller and
  Sch{\"{u}}tze}2011]{mueller2011improved}
Thomas M{\"{u}}ller and Hinrich Sch{\"{u}}tze.
\newblock 2011.
\newblock Improved modeling of out-of-vocabulary words using morphological
  classes.
\newblock In {\em Proceedings of the 49th Annual Meeting of the Association for
  Computational Linguistics: Human Language Technologies}, pages 524--528,
  Portland, Oregon, USA, June. Association for Computational Linguistics.

\bibitem[\protect\citename{Peters \bgroup et al.\egroup }2018]{elmo}
Matthew Peters, Mark Neumann, Mohit Iyyer, Matt Gardner, Christopher Clark,
  Kenton Lee, and Luke Zettlemoyer.
\newblock 2018.
\newblock Deep contextualized word representations.
\newblock In {\em Proceedings of the 2018 Conference of the North {A}merican
  Chapter of the Association for Computational Linguistics: Human Language
  Technologies, Volume 1 (Long Papers)}, pages 2227--2237, New Orleans,
  Louisiana, June. Association for Computational Linguistics.

\bibitem[\protect\citename{Pinter \bgroup et al.\egroup }2017]{mimick}
Yuval Pinter, Robert Guthrie, and Jacob Eisenstein.
\newblock 2017.
\newblock Mimicking word embeddings using subword {RNN}s.
\newblock In {\em Proceedings of the 2017 Conference on Empirical Methods in
  Natural Language Processing}, pages 102--112, Copenhagen, Denmark, September.
  Association for Computational Linguistics.

\bibitem[\protect\citename{Pinter \bgroup et al.\egroup }2020]{unblend}
Yuval Pinter, Cassandra~L. Jacobs, and Jacob Eisenstein.
\newblock 2020.
\newblock Will it unblend?
\newblock In {\em Findings of EMNLP}.

\bibitem[\protect\citename{Sakaguchi \bgroup et al.\egroup
  }2017]{sakaguchi2017robsut}
Keisuke Sakaguchi, Kevin Duh, Matt Post, and Benjamin Van~Durme.
\newblock 2017.
\newblock Robsut wrod reocginiton via semi-character recurrent neural network.
\newblock In {\em Thirty-First AAAI Conference on Artificial Intelligence}.

\bibitem[\protect\citename{van~der Goot}2019]{vandergoot2019depth}
Rob van~der Goot.
\newblock 2019.
\newblock An in-depth analysis of the effect of lexical normalization on the
  dependency parsing of social media.
\newblock In {\em Proceedings of the 5th Workshop on Noisy User-generated Text
  (W-NUT 2019)}, pages 115--120, Hong Kong, China, November. Association for
  Computational Linguistics.

\bibitem[\protect\citename{Wales and Ramsaran}1990]{wales1990phonotactics}
Katie Wales and S~Ramsaran.
\newblock 1990.
\newblock Phonotactics and phonaesthesia: the power of folk lexicology.
\newblock {\em Studies in pronunciation of English. A commemorative volume in
  honour of AC Gimson}, pages 339--351.

\bibitem[\protect\citename{Young \bgroup et al.\egroup }2018]{young2018recent}
Tom Young, Devamanyu Hazarika, Soujanya Poria, and Erik Cambria.
\newblock 2018.
\newblock Recent trends in deep learning based natural language processing.
\newblock {\em ieee Computational intelligenCe magazine}, 13(3):55--75.

\bibitem[\protect\citename{Zwicky}2010]{zwicky2010libfixes}
Arnold Zwicky.
\newblock 2010.
\newblock Libfixes.
\newblock {\em Arnold Zwicky's Blog}.

\end{thebibliography}

\end{document}